\pdfoutput=1
\RequirePackage{fix-cm}
\documentclass[10pt,letterpaper]{article}
\usepackage[utf8]{inputenc}
\usepackage[T1]{fontenc}
\usepackage[english]{babel}
\usepackage{sourcesanspro}
\usepackage{amsmath,amssymb}
\usepackage{microtype}
\usepackage[dvipsnames]{xcolor}
\usepackage{graphicx,booktabs,tabularx,array,fancyhdr,float}
\usepackage[numbers,sort&compress]{natbib}
\usepackage[explicit]{titlesec}
\usepackage[skins]{tcolorbox}
\usepackage{caption}
\usepackage{placeins,needspace}
\usepackage[left=2.5cm,right=2.5cm,top=2.5cm,bottom=2.5cm,footskip=30pt]{geometry}
\definecolor{ReportAccent}{HTML}{B04F1A}
\definecolor{ReportLight}{HTML}{F5D8BF}
\definecolor{ReportAbstractBackground}{HTML}{FFF1E5}

\usepackage[colorlinks=true,allcolors=ReportAccent]{hyperref}
\hypersetup{pdftitle={Predictive Long-Horizon Progress Modeling with Action-Aware Quantification},pdfauthor={Tianchi Xu, Yalin Zheng, Yangxin Zhu, Ruonan Ren, Zheng Fu, Yunlong Wang, Yuanxin Zhong}}
\newcommand{\ReportTitle}{Predictive Long-Horizon Progress Modeling\\with Action-Aware Quantification}
\newcommand{\ReportAuthor}{\mbox{\href{mailto:tianchixu@stu.pku.edu.cn}{\color{black}Tianchi Xu}\textsuperscript{1,3,*}}\hspace{4.5pt}\mbox{\href{mailto:ayleea.zheng@rimbot.com}{\color{black}Yalin Zheng}\textsuperscript{1,*}}\hspace{4.5pt}\mbox{\href{mailto:zhuyangxin@rimbot.com}{\color{black}Yangxin Zhu}\textsuperscript{1}}\hspace{4.5pt}\mbox{\href{mailto:ruonan.ren@rimbot.com}{\color{black}Ruonan Ren}\textsuperscript{1}}\hspace{4.5pt}\mbox{\href{mailto:fuzheng039@gmail.com}{\color{black}Zheng Fu}\textsuperscript{1,2}}\hspace{4.5pt}\mbox{\href{mailto:yl-wang19@tsinghua.org.cn}{\color{black}Yunlong Wang}\textsuperscript{1}}\hspace{4.5pt}\mbox{\href{mailto:jacob@rimbot.com}{\color{black}Yuanxin Zhong}\textsuperscript{1,\ensuremath{\dagger}}}}
\newcommand{\ReportAffiliation}{\textsuperscript{1}\,RIMBOT\qquad\textsuperscript{2}\,Tsinghua University\qquad\textsuperscript{3}\,Peking University}
\newcommand{\ReportEmail}{jacob@rimbot.com}
\newcommand{\ReportDate}{October 23, 2026}
\newcommand{\ReportProjectPage}{\href{https://rimbot.com/en/research/nimcritic/}{\texttt{https://rimbot.com/en/research/nimcritic/}}}
\fancypagestyle{plain}{\fancyhf{}\fancyfoot[C]{\normalfont\fontsize{10}{12}\selectfont\thepage}}
\titleformat{\section}{\sffamily\bfseries\fontsize{12}{14}\selectfont\color{black}}{\thesection}{0.75em}{#1}
\titleformat{\subsection}{\sffamily\bfseries\fontsize{11}{13}\selectfont\color{black}}{\thesubsection}{0.65em}{#1}
\titleformat{\subsubsection}{\sffamily\bfseries\fontsize{10}{12}\selectfont}{\thesubsubsection}{0.5em}{#1}
\titlespacing*{\section}{0pt}{12pt plus2pt minus1pt}{5pt}
\titlespacing*{\subsection}{0pt}{10pt plus2pt minus1pt}{4pt}
\titlespacing*{\paragraph}{0pt}{8pt}{0.55em}
\newcommand{\WRMMetric}[1]{\makebox[\linewidth][c]{{\sffamily\fontsize{7.5}{9}\selectfont #1}}}
\newcommand{\WRMGroup}[1]{{\sffamily\fontsize{8}{9.5}\selectfont #1}}


\newcommand{\ReportAbstract}{%
Quantitative and dense task progress feedback is essential for robotic manipulation across data curation, policy evaluation, and reinforcement learning. Existing progress annotations often rely on sparse keyframe markings derived from subjective manual judgments, limiting their scalability and cross-task consistency, particularly in long-horizon manipulation. To overcome these challenges, we propose \textbf{action-aware quantification}, a unified progress specification that decomposes long-horizon trajectories into object-level subtasks via a vision--language model (VLM) and assigns weighted action credits to capture fine-grained execution dynamics. Using VLM auto-labeling and human calibration, we construct an 800+ hour training video dataset and \textbf{LOOM}, a long-horizon progress benchmark containing 19,618 human-calibrated evaluation frames. To model these dense progress dynamics, we introduce \textbf{NimCritic}, a predictive progress critic trained via Joint Video--Progress Learning. We further incorporate Historical Progress Perturbation to mitigate sequential drift, and design a contrastive Video--Progress Alignment objective that bridges asymmetric expert streams by mutually grounding intermediate visual dynamics and denoised progress trajectories. \textbf{NimCritic} matches or outperforms state-of-the-art process reward models (PRMs) on the established Video Progress Benchmark (VPB) and \textbf{LOOM}, surpassing GVL (GPT-6-Sol) on VPB by about 65\% in terminal success accuracy (TSA). On the large-scale RoboArena policy evaluation benchmark, it also retrieves human-selected successful rollouts with about 97\% overall recall. These results establish \textbf{action-aware quantification} as a scalable foundation for dense progress feedback in robotic manipulation.
}

\renewcommand{\maketitle}{%
\begingroup
\setlength{\parskip}{0pt}
\vspace*{-44pt}
\noindent\begin{minipage}[c]{0.58\linewidth}
\includegraphics[height=6.0mm]{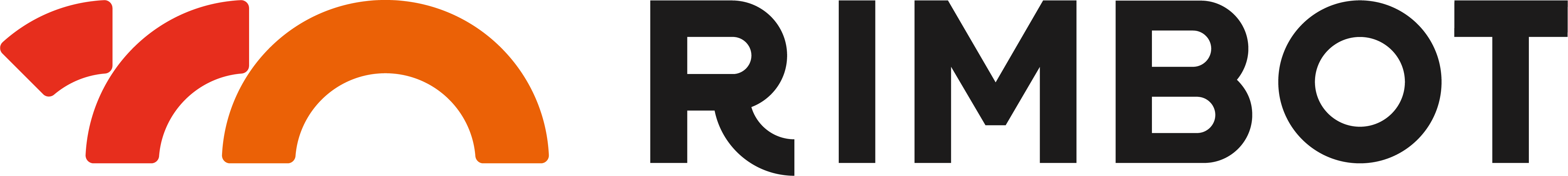}
\end{minipage}\hfill
\begin{minipage}[c]{0.38\linewidth}\raggedleft
\includegraphics[height=9mm]{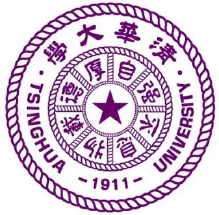}\hspace{4mm}%
\includegraphics[height=9mm]{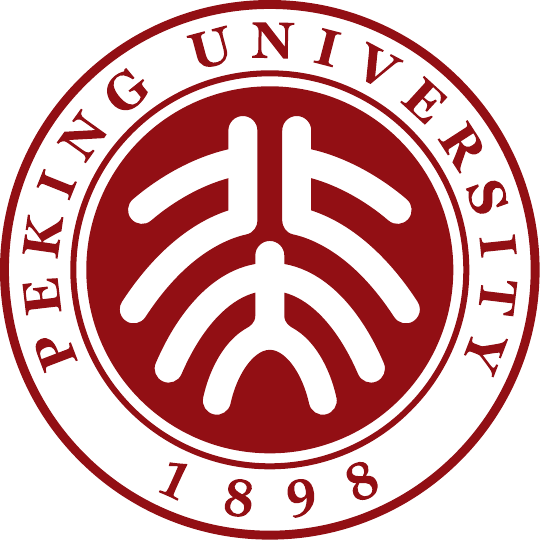}
\end{minipage}\par
\vspace{2pt}
{\color{ReportAccent}\rule{\linewidth}{0.5pt}\par}
\vspace{2pt}
{\centering\sffamily\bfseries\fontsize{17}{20}\selectfont\ReportTitle\par}
\vspace{2pt}
{\color{ReportAccent}\rule{\linewidth}{0.5pt}\par}
\vspace{2.5pt}
{\centering\sffamily\bfseries\fontsize{9.8}{11.8}\selectfont\ReportAuthor\par}
\vspace{1.5pt}
{\centering\sffamily\fontsize{10}{12}\selectfont\ReportAffiliation\par}
\vspace{1.5pt}
{\centering\normalfont\fontsize{9}{11}\selectfont\textsuperscript{*}Equal contribution.\qquad\textsuperscript{\ensuremath{\dagger}}Corresponding author.\par}
\vspace{2pt}
\begin{tcolorbox}[enhanced,colback=ReportAbstractBackground,colframe=ReportLight,boxrule=0.6pt,
arc=7pt,outer arc=7pt,boxsep=0pt,left=20pt,right=20pt,top=4.5pt,bottom=4.5pt,
before skip=0pt,after skip=0pt]
{\centering\sffamily\bfseries\fontsize{11.5}{13.5}\selectfont\color{black}Abstract\par}
\vspace{2pt}
{\normalfont\fontsize{9.0}{10.6}\selectfont\ReportAbstract\par}
\vspace{2pt}
{\normalfont\fontsize{8.8}{10.5}\selectfont
\textsf{\textbf{Date:}}\quad\ReportDate\par
\textsf{\textbf{Correspondence:}}\quad\href{mailto:\ReportEmail}{\texttt{\ReportEmail}}\par
\textsf{\textbf{Project Page:}}\quad\ReportProjectPage\par}
\end{tcolorbox}
\vspace{2pt}
\enlargethispage{28pt}
\noindent\begin{minipage}{\linewidth}
\centering
\includegraphics[width=\linewidth]{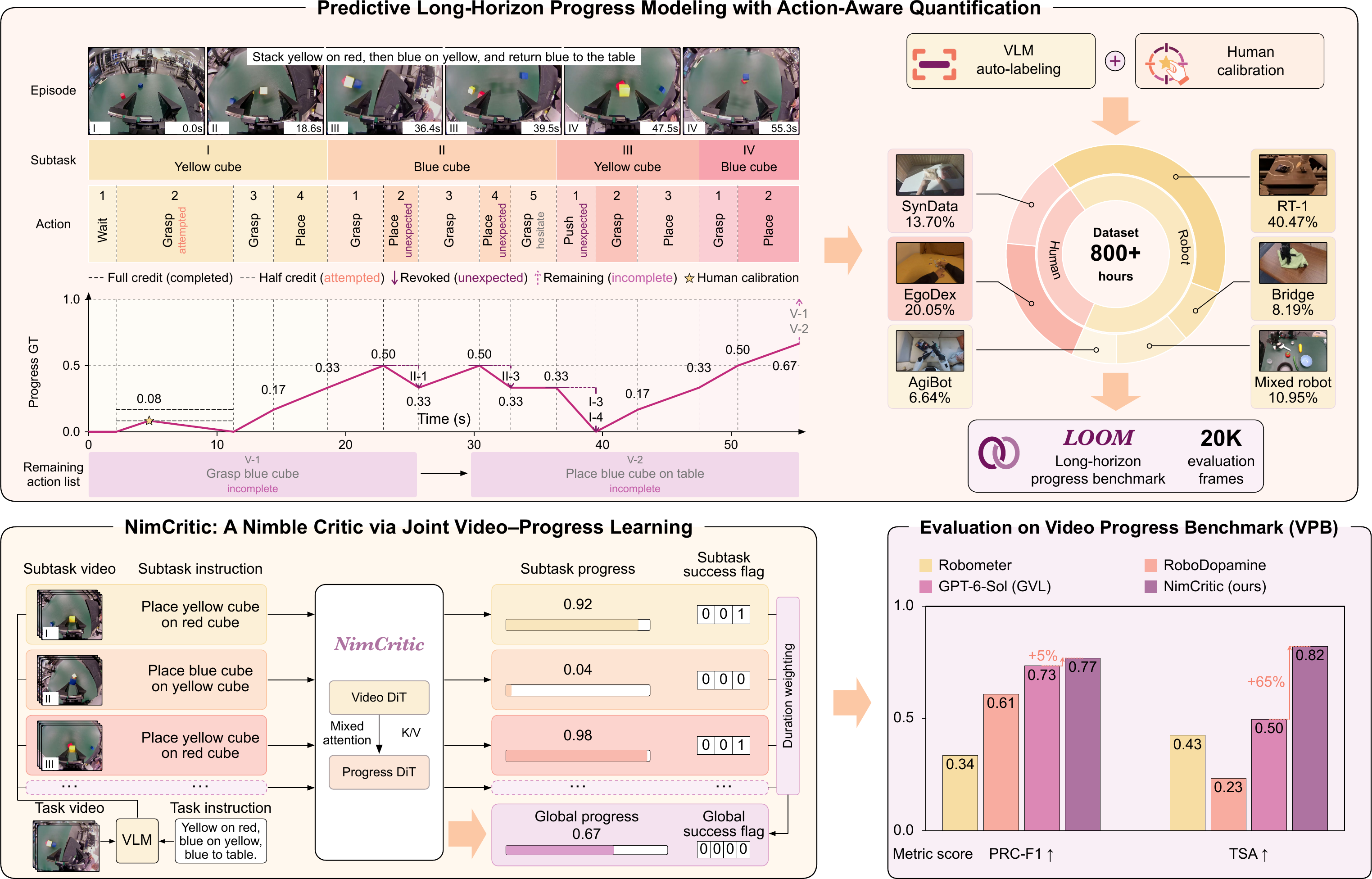}\par
\captionsetup{font=small,skip=2.5pt}
\captionof{figure}{\textbf{Overview of action-aware quantification and NimCritic:} cube stacking illustrates earned credit, temporary partial credit for an unsuccessful grasp, revocation, and recovery; the lower panels show subtask progress composition and VPB performance.}
\label{fig:overview}\par
\end{minipage}
\endgroup
\clearpage
}

\newcommand{\WRMTableBenchmark}{%
\begin{table}[!htbp]
\centering
\begingroup
\fontsize{8.0}{9.7}\selectfont
\setlength{\tabcolsep}{0.25pt}
\renewcommand{\arraystretch}{1.15}
\begin{tabularx}{\linewidth}{@{}>{\fontsize{7.7}{9.3}\selectfont}l*{12}{>{\centering\arraybackslash}X}@{}}
\toprule
Method & \multicolumn{4}{c}{\WRMGroup{Overall}} & \multicolumn{4}{c}{\WRMGroup{Single Task}} & \multicolumn{4}{c}{\WRMGroup{Multi Task}} \\
\cmidrule(lr){2-5}\cmidrule(lr){6-9}\cmidrule(l){10-13}
 & \WRMMetric{PRC$^+\!\uparrow$} & \WRMMetric{PRC$^-\!\uparrow$} & \WRMMetric{PRC-F1$\uparrow$} & \WRMMetric{TSA$\uparrow$} & \WRMMetric{PRC$^+\!\uparrow$} & \WRMMetric{PRC$^-\!\uparrow$} & \WRMMetric{PRC-F1$\uparrow$} & \WRMMetric{TSA$\uparrow$} & \WRMMetric{PRC$^+\!\uparrow$} & \WRMMetric{PRC$^-\!\uparrow$} & \WRMMetric{PRC-F1$\uparrow$} & \WRMMetric{TSA$\uparrow$} \\
\midrule
\multicolumn{13}{@{}l}{\emph{General-Purpose VLMs with GVL}} \\
GVL (Gemini 3.8 Flash) & \textbf{0.73} & \textbf{0.73} & \textbf{0.73} & 0.60 & \textbf{0.69} & \textbf{0.69} & \textbf{0.69} & 0.57 & \textbf{0.74} & \textbf{0.74} & \textbf{0.74} & 0.61 \\
GVL (GPT-6-Sol) & 0.68 & 0.68 & 0.68 & \textbf{0.65} & 0.69 & 0.69 & 0.69 & \textbf{0.65} & 0.68 & 0.68 & 0.68 & \textbf{0.65} \\
\midrule
\multicolumn{13}{@{}l}{\emph{Specialized Reward Models}} \\
ProcVLM 2B & 0.48 & 0.19 & 0.27 & 0.26 & 0.57 & 0.20 & 0.29 & 0.52 & 0.44 & 0.19 & 0.26 & 0.13 \\
VLAC-CUT & 0.84 & -0.10 & 0.00 & 0.60 & \textbf{0.80} & 0.42 & 0.55 & 0.51 & 0.80 & -0.65 & 0.00 & 0.71 \\
LRM & 0.15 & 0.15 & 0.15 & 0.23 & 0.26 & 0.26 & 0.26 & 0.41 & 0.10 & 0.10 & 0.10 & 0.14 \\
Robometer 4B & 0.54 & 0.11 & 0.18 & 0.56 & 0.58 & 0.35 & 0.44 & 0.67 & 0.52 & -0.01 & 0.00 & 0.51 \\
RynnValue 4B & 0.17 & 0.55 & 0.26 & 0.28 & 0.33 & 0.50 & 0.40 & 0.31 & 0.10 & 0.58 & 0.17 & 0.26 \\
VLAC-Critic 2B & 0.30 & 0.01 & 0.02 & 0.15 & 0.45 & -0.16 & 0.00 & 0.25 & 0.24 & 0.08 & 0.12 & 0.10 \\
VLAC-Critic 8B & 0.59 & 0.11 & 0.18 & 0.19 & 0.59 & 0.24 & 0.34 & 0.31 & 0.60 & 0.03 & 0.07 & 0.13 \\
RoboDopamine-GRM-2 8B & 0.38 & 0.64 & 0.48 & 0.29 & 0.54 & 0.62 & 0.58 & 0.35 & 0.31 & 0.65 & 0.42 & 0.27 \\
\textbf{NimCritic} & \textbf{0.87} & \textbf{0.86} & \textbf{0.86} & \textbf{0.92} & 0.70 & \textbf{0.64} & \textbf{0.67} & \textbf{0.88} & \textbf{0.95} & \textbf{0.95} & \textbf{0.95} & \textbf{0.93} \\
\bottomrule
\end{tabularx}
\endgroup
\caption{\textbf{Progress estimation on LOOM.}}
\label{tab:benchmark}
\end{table}%
}

\newcommand{\WRMTableVPB}{%
\begin{table}[H]
\centering
\begingroup
\fontsize{8.0}{9.7}\selectfont
\setlength{\tabcolsep}{1.0pt}
\renewcommand{\arraystretch}{1.15}
\begin{tabularx}{\linewidth}{@{}>{\fontsize{7.7}{9.3}\selectfont}l*{10}{>{\centering\arraybackslash}X}@{}}
\toprule
Method & \multicolumn{2}{c}{\WRMGroup{Overall}} & \multicolumn{2}{c}{\WRMGroup{Expert Seen}} & \multicolumn{2}{c}{\WRMGroup{Expert Unseen}} & \multicolumn{2}{c}{\WRMGroup{Non-expert Seen}} & \multicolumn{2}{c}{\WRMGroup{Non-expert Unseen}} \\
\cmidrule(lr){2-3}\cmidrule(lr){4-5}\cmidrule(lr){6-7}\cmidrule(lr){8-9}\cmidrule(l){10-11}
 & \WRMMetric{PRC-F1$\uparrow$} & \WRMMetric{TSA$\uparrow$} & \WRMMetric{PRC-F1$\uparrow$} & \WRMMetric{TSA$\uparrow$} & \WRMMetric{PRC-F1$\uparrow$} & \WRMMetric{TSA$\uparrow$} & \WRMMetric{PRC-F1$\uparrow$} & \WRMMetric{TSA$\uparrow$} & \WRMMetric{PRC-F1$\uparrow$} & \WRMMetric{TSA$\uparrow$} \\
\midrule
\multicolumn{11}{@{}l}{\emph{General-Purpose VLMs with GVL}} \\
GVL (Gemini 3.8 Flash) & 0.73 & 0.47 & \textbf{0.79} & 0.43 & \textbf{0.77} & \textbf{0.39} & 0.66 & 0.54 & 0.66 & 0.56 \\
GVL (GPT-6-Sol) & \textbf{0.73} & \textbf{0.50} & 0.77 & \textbf{0.43} & 0.75 & 0.39 & \textbf{0.71} & \textbf{0.63} & \textbf{0.69} & \textbf{0.58} \\
\midrule
\multicolumn{11}{@{}l}{\emph{Specialized Reward Models}} \\
ProcVLM 2B & 0.00 & 0.36 & 0.00 & 0.27 & 0.00 & 0.26 & 0.15 & 0.46 & 0.23 & 0.48 \\
VLAC-CUT & 0.00 & 0.67 & 0.00 & 0.68 & 0.00 & 0.64 & 0.40 & 0.78 & 0.00 & 0.76 \\
LRM & 0.26 & 0.21 & 0.28 & 0.10 & 0.23 & 0.08 & 0.28 & 0.35 & 0.27 & 0.37 \\
Robometer 4B & 0.34 & 0.43 & 0.36 & 0.37 & 0.26 & 0.30 & 0.34 & 0.53 & 0.38 & 0.55 \\
RynnValue 4B & 0.50 & 0.15 & 0.56 & 0.04 & 0.49 & 0.06 & 0.46 & 0.28 & 0.46 & 0.26 \\
VLAC-Critic 2B & 0.00 & 0.11 & 0.17 & 0.01 & 0.11 & 0.02 & 0.00 & 0.24 & 0.00 & 0.22 \\
VLAC-Critic 8B & 0.50 & 0.15 & 0.63 & 0.06 & 0.61 & 0.06 & 0.32 & 0.26 & 0.36 & 0.26 \\
RoboDopamine-GRM-2 8B & 0.61 & 0.23 & 0.69 & 0.14 & 0.64 & 0.14 & 0.52 & 0.35 & 0.54 & 0.34 \\
\textbf{NimCritic} & \textbf{0.77} & \textbf{0.82} & \textbf{0.88} & \textbf{0.86} & \textbf{0.83} & \textbf{0.84} & \textbf{0.66} & \textbf{0.80} & \textbf{0.67} & \textbf{0.78} \\
\bottomrule
\end{tabularx}
\endgroup
\caption{\textbf{Progress estimation on VPB.}}
\label{tab:vpb}
\end{table}%
}

\newcommand{\WRMTableData}{%
\begin{table}[H]
\centering
\begingroup
\fontsize{8.5}{10.2}\selectfont
\setlength{\tabcolsep}{2.5pt}
\renewcommand{\arraystretch}{1.15}
\begin{tabularx}{0.82\linewidth}{@{}*{3}{>{\centering\arraybackslash}p{0.115\linewidth}}*{4}{>{\centering\arraybackslash}X}@{}}
\toprule
\multicolumn{3}{c}{Additional Training Data} & \multicolumn{4}{c}{VPB (Overall)} \\
\cmidrule(lr){1-3}\cmidrule(l){4-7}
AgiBot & Egocentric & Non-Expert & \WRMMetric{PRC$^+\!\uparrow$} & \WRMMetric{PRC$^-\!\uparrow$} & \WRMMetric{PRC-F1$\uparrow$} & \WRMMetric{TSA$\uparrow$} \\
\midrule
$\times$ & $\times$ & $\times$ & 0.62 & 0.42 & 0.50 & 0.42 \\
$\times$ & $\checkmark$ & $\times$ & 0.66 & 0.72 & 0.69 & 0.86 \\
$\times$ & $\times$ & $\checkmark$ & 0.72 & 0.56 & 0.63 & \textbf{0.88} \\
$\checkmark$ & $\checkmark$ & $\checkmark$ & \textbf{0.73} & \textbf{0.81} & \textbf{0.77} & 0.82 \\
\bottomrule
\end{tabularx}
\endgroup
\caption{\textbf{Evaluation of training data compositions on VPB.}}
\label{tab:data-study}
\end{table}%
}

\newcommand{\WRMTableComponents}{%
\begin{table}[H]
\centering
\begingroup
\fontsize{8.0}{9.6}\selectfont
\setlength{\tabcolsep}{2.0pt}
\renewcommand{\arraystretch}{1.15}
\begin{tabularx}{0.82\linewidth}{@{}*{3}{>{\centering\arraybackslash}p{0.155\linewidth}}*{4}{>{\centering\arraybackslash}X}@{}}
\toprule
\multicolumn{3}{c}{Design Components} & \multicolumn{4}{c}{VPB (Overall)} \\
\cmidrule(lr){1-3}\cmidrule(l){4-7}
\shortstack{Joint Video--\\Progress Learning} & \shortstack{Historical Progress\\Perturbation} & \shortstack{Video--Progress\\Alignment} & \WRMMetric{PRC$^+\!\uparrow$} & \WRMMetric{PRC$^-\!\uparrow$} & \WRMMetric{PRC-F1$\uparrow$} & \WRMMetric{TSA$\uparrow$} \\
\midrule
$\times$ & $\times$ & $\times$ & 0.68 & 0.24 & 0.35 & 0.86 \\
$\checkmark$ & $\times$ & $\times$ & 0.68 & 0.53 & 0.59 & 0.75 \\
$\checkmark$ & $\checkmark$ & $\times$ & \textbf{0.74} & 0.60 & 0.67 & \textbf{0.87} \\
$\checkmark$ & $\checkmark$ & $\checkmark$ & 0.73 & \textbf{0.81} & \textbf{0.77} & 0.82 \\
\bottomrule
\end{tabularx}
\endgroup
\caption{\textbf{Ablation study on VPB.}}
\label{tab:components}
\end{table}%
}

\newcommand{\WRMTablePolicy}{%
\begin{table}[H]
\centering
\begingroup
\fontsize{8.0}{9.7}\selectfont
\setlength{\tabcolsep}{2.3pt}
\renewcommand{\arraystretch}{1.15}
\begin{tabularx}{\linewidth}{@{}>{\fontsize{7.7}{9.3}\selectfont}l*{8}{>{\centering\arraybackslash}X}@{}}
\toprule
Method & \multicolumn{2}{c}{2 policies} & \multicolumn{2}{c}{4 policies} & \multicolumn{2}{c}{7 policies} & \multicolumn{2}{c}{Overall} \\
\cmidrule(lr){2-3}\cmidrule(lr){4-5}\cmidrule(lr){6-7}\cmidrule(l){8-9}
 & R (\%)$\uparrow$ & HR (\%)$\uparrow$ & R (\%)$\uparrow$ & HR (\%)$\uparrow$ & R (\%)$\uparrow$ & HR (\%)$\uparrow$ & R (\%)$\uparrow$ & HR (\%)$\uparrow$ \\
\midrule
ProcVLM 2B & 50.00 & 52.58 & 21.43 & 37.50 & 20.57 & 28.86 & 36.89 & 45.45 \\
VLAC-CUT & 45.27 & 48.97 & 42.86 & 37.50 & 63.21 & 69.92 & 53.03 & 55.02 \\
LRM & 26.78 & 26.46 & 7.14 & 12.50 & 35.09 & 34.96 & 30.16 & 28.83 \\
Robometer 4B & 50.15 & 51.20 & 28.57 & 25.00 & 62.08 & 62.60 & 55.08 & 54.31 \\
RynnValue 4B & 72.49 & 72.51 & 21.43 & 12.50 & 48.68 & 56.91 & 61.56 & 67.34 \\
VLAC-Critic 2B & 12.28 & 12.89 & 0.00 & 0.00 & 3.02 & 5.28 & 8.11 & 10.53 \\
VLAC-Critic 8B & 19.23 & 19.59 & 0.00 & 0.00 & 7.74 & 12.20 & 14.02 & 17.22 \\
RoboDopamine-GRM-2 8B & 93.34 & 94.16 & 78.57 & \textbf{87.50} & 87.55 & 92.68 & 90.66 & 93.66 \\
\textbf{NimCritic} & \textbf{98.08} & \textbf{98.45} & \textbf{85.71} & \textbf{87.50} & \textbf{95.28} & \textbf{97.56} & \textbf{96.72} & \textbf{98.09} \\
\bottomrule
\end{tabularx}
\endgroup
\caption{\textbf{Policy rollout evaluation on RoboArena.}}
\label{tab:policy}
\end{table}%
}

\newcommand{\WRMTableSegmenters}{%
\begin{table}[!htbp]
\centering
\begingroup
\fontsize{8.5}{10.2}\selectfont
\setlength{\tabcolsep}{3pt}
\renewcommand{\arraystretch}{1.15}
\begin{tabularx}{\linewidth}{@{}l*{4}{>{\centering\arraybackslash}X}@{}}
\toprule
Segmenter & \WRMMetric{SegCover-F1$\uparrow$} & \WRMMetric{Boundary F1$\uparrow$} & \WRMMetric{ARI$\uparrow$} & \WRMMetric{SegCover-F1 (1:1)$\uparrow$} \\
\midrule
GPT-6-Sol & 0.78 & 0.35 & 0.66 & 0.76 \\
Qwen 3.8 Max & 0.79 & 0.40 & \textbf{0.70} & 0.77 \\
Gemini 3.8 Flash & \textbf{0.80} & \textbf{0.41} & 0.69 & \textbf{0.78} \\
\bottomrule
\end{tabularx}
\endgroup
\caption{\textbf{VLM-based subtask decomposition on LOOM.}}
\label{tab:segmenters}
\end{table}%
}

\newcommand{\WRMTableCommonSegmentation}{%
\begin{table}[!htbp]
\centering
\begingroup
\fontsize{8.5}{10.2}\selectfont
\setlength{\tabcolsep}{3pt}
\renewcommand{\arraystretch}{1.15}
\begin{tabularx}{\linewidth}{@{}l*{4}{>{\centering\arraybackslash}X}@{}}
\toprule
Method & \WRMMetric{PRC$^+\!\uparrow$} & \WRMMetric{PRC$^-\!\uparrow$} & \WRMMetric{PRC-F1$\uparrow$} & \WRMMetric{TSA$\uparrow$} \\
\midrule
\multicolumn{5}{@{}l}{\emph{General-Purpose VLMs with GVL}} \\
GVL (Gemini 3.8 Flash) & 0.92 & 0.92 & 0.92 & 0.61 \\
GVL (GPT-6-Sol) & \textbf{0.96} & \textbf{0.96} & \textbf{0.96} & \textbf{0.65} \\
\midrule
\multicolumn{5}{@{}l}{\emph{Specialized Reward Models}} \\
ProcVLM 2B & 0.87 & 0.54 & 0.67 & 0.13 \\
VLAC-CUT & 0.82 & 0.37 & 0.51 & 0.71 \\
LRM & 0.68 & 0.68 & 0.68 & 0.14 \\
Robometer 4B & 0.89 & 0.70 & 0.79 & 0.51 \\
RynnValue 4B & 0.85 & 0.88 & 0.86 & 0.26 \\
VLAC-Critic 2B & 0.43 & 0.01 & 0.01 & 0.10 \\
VLAC-Critic 8B & 0.62 & -0.22 & 0.00 & 0.13 \\
RoboDopamine-GRM-2 8B & 0.63 & 0.87 & 0.73 & 0.27 \\
\textbf{NimCritic} & \textbf{0.95} & \textbf{0.95} & \textbf{0.95} & \textbf{0.93} \\
\bottomrule
\end{tabularx}
\endgroup
\caption{\textbf{Progress estimation with shared Gemini 3.8 Flash decomposition on LOOM.}}
\label{tab:common-segmentation}
\end{table}%
}

\newcommand{\WRMTableDirectVsFull}{%
\begin{table}[!htbp]
\centering
\begingroup
\fontsize{8.0}{9.7}\selectfont
\setlength{\tabcolsep}{0.25pt}
\renewcommand{\arraystretch}{1.15}
\begin{tabularx}{\linewidth}{@{}>{\fontsize{7.7}{9.3}\selectfont}l*{12}{>{\centering\arraybackslash}X}@{}}
\toprule
Method & \multicolumn{4}{c}{\WRMGroup{Overall}} & \multicolumn{4}{c}{\WRMGroup{Single Task}} & \multicolumn{4}{c}{\WRMGroup{Multi Task}} \\
\cmidrule(lr){2-5}\cmidrule(lr){6-9}\cmidrule(l){10-13}
 & \WRMMetric{PRC$^+\!\uparrow$} & \WRMMetric{PRC$^-\!\uparrow$} & \WRMMetric{PRC-F1$\uparrow$} & \WRMMetric{TSA$\uparrow$} & \WRMMetric{PRC$^+\!\uparrow$} & \WRMMetric{PRC$^-\!\uparrow$} & \WRMMetric{PRC-F1$\uparrow$} & \WRMMetric{TSA$\uparrow$} & \WRMMetric{PRC$^+\!\uparrow$} & \WRMMetric{PRC$^-\!\uparrow$} & \WRMMetric{PRC-F1$\uparrow$} & \WRMMetric{TSA$\uparrow$} \\
\midrule
\shortstack[l]{\textbf{NimCritic w/o}\\\textbf{Subtask Decomposition}} & 0.64 & 0.54 & 0.59 & \textbf{0.92} & \textbf{0.70} & \textbf{0.64} & \textbf{0.67} & \textbf{0.88} & 0.61 & 0.49 & 0.55 & \textbf{0.93} \\
\textbf{NimCritic} & \textbf{0.87} & \textbf{0.86} & \textbf{0.86} & \textbf{0.92} & \textbf{0.70} & \textbf{0.64} & \textbf{0.67} & \textbf{0.88} & \textbf{0.95} & \textbf{0.95} & \textbf{0.95} & \textbf{0.93} \\
\bottomrule
\end{tabularx}
\endgroup
\caption{\textbf{Direct and subtask-based scoring with the same NimCritic on LOOM.}}
\label{tab:direct-vs-full}
\end{table}%
}

\begin{document}
\maketitle
\section{Introduction}
\label{sec:introduction}
\enlargethispage{2\baselineskip}
Process reward models (PRMs) evaluate intermediate execution states from visual observations, supplying dense feedback for robotic data curation, policy evaluation, and reinforcement learning~\cite{robo_dopamine,vlac}.
Consider a robot instructed to stack blocks in sequence: a completed placement advances the task, an attempted grasp earns temporary partial credit, and a dislodged block triggers revocation of its earlier placement credit.
Faithful process supervision must track these state changes across hesitation, progress regression, and recovery.
Recent methods draw on vision--language model (VLM)-based reward learning to bring task semantics to visual evaluation, and on video world models to supervise temporal dynamics through future prediction~\cite{liv,wvm}.

Current PRMs learn relative progress through trajectory comparisons~\cite{robometer}, interpolate across manually annotated task stages~\cite{sarm}, or assign VLM-generated subtask budgets~\cite{procvlm}.
Relative comparisons establish trajectory ordering while leaving cross-task progress magnitude unspecified, and elapsed-time interpolation varies with execution speed for the same achieved state~\cite{sarm}.
Long-horizon feedback must also quantify what a partial attempt contributes and which credit is revoked when a later action destroys an earlier achievement.
Scaling process evaluation thus requires both a unified absolute progress specification and a critic that tracks non-monotonic execution across subtasks.

To overcome these challenges, we propose \textbf{action-aware quantification}. We first define a subtask as a sequence of operations on the same manipulated object.
We then allocate a fixed local budget using the weights of executed actions and the necessary remaining actions. Action boundaries locate the curve breakpoints, and retained weighted action credit determines local progress within this budget.
Forward effects receive full credit, engaged-only attempts receive temporary half credit, and subsequent revocations remove the credit of the referenced earlier occurrence.
Finally, global progress tracks retained contributions across subtasks, including credit revoked by later destructive actions.
VLM pre-annotation and human verification establish structured action records, from which rule-based post-processing computes dense progress for training and evaluation.
We assemble a training dataset of 253,276 episodes spanning 800+ hours of robot and human video, together with \textbf{LOOM} (a \emph{LOng-hOrizon Manipulation progress benchmark}), comprising 300 cross-embodiment trajectories evaluated at 19,618 human-calibrated frames.
The benchmark follows successive subtasks in long-horizon execution, covering different rates of advancement, hesitation, regression, recovery, and incomplete execution.

Building on this foundation, we introduce \textbf{NimCritic} (\emph{A Nimble Critic via Joint Video--Progress Learning}), a dual-diffusion-transformer world reward model with separately supervised progress and completion outputs.
We term the model \emph{nimble} because it uses predictive video learning during training and scores subtasks in parallel at inference~\cite{yuan2026fastwam}, and a \emph{critic} because it evaluates task advancement and goal satisfaction.
Joint Video--Progress Learning models how the observed state evolves while tying those transitions to advancement toward the specified subtask goal, complemented by Historical Progress Perturbation to regularize sequential history conditioning and contrastive Video--Progress Alignment to strengthen cross-modal correspondence.
On the established Video Progress Benchmark (VPB), \textbf{NimCritic} achieves 0.77 overall PRC-F1 and 0.82 TSA (surpassing GVL with GPT-6-Sol by about 65\%).
On LOOM, it leads all external baselines across all four metrics in the overall and multi-task groups (0.86 and 0.95 PRC-F1), and on RoboArena it retrieves human-selected successful rollouts with 96.72\% overall recall and 98.09\% hit rate.

We summarize our main contributions as follows:
\begin{itemize}\setlength{\itemsep}{1pt}\setlength{\parskip}{0pt}\setlength{\parsep}{0pt}
\item[(1)] We formulate \textbf{action-aware quantification}, which converts object-level action effects into dense progress through weighted credits, explicit revocation, and composition across complete and incomplete subtasks.
\item[(2)] We construct an 800+ hour action-aware training dataset through VLM pre-annotation and human verification, and \textbf{LOOM}, a 300-trajectory benchmark with 19,618 human-calibrated evaluation frames for testing long-horizon progress through hesitation, regression, recovery, and incomplete execution.
\item[(3)] We propose \textbf{NimCritic}, which combines predictive video learning with progress and completion supervision, using Historical Progress Perturbation and Video--Progress Alignment to support sequential estimates and connect visual features to task progress.
\item[(4)] We demonstrate the \textbf{highest} overall PRC-F1 and TSA on VPB and LOOM, together with 96.72\% successful-rollout recall on RoboArena, and analyze the roles of egocentric human videos, non-expert trajectories, and predictive supervision.
\end{itemize}

\clearpage
\section{Related Work}
\label{sec:related-work}
\paragraph{Robotic Process Reward Models.}
Robotic PRMs map visual observations and task instructions to dense intermediate evaluation signals.
One line of research builds on vision--language representations: language--image alignment provides goal-conditioned rewards~\cite{liv}, while training-free methods infer progress from shuffled frame ordering or completion-token probabilities~\cite{gvl,topreward}.
Learned VLM critics further supervise relative progress via history-conditioned trajectory comparisons or signed temporal-warp objectives~\cite{robo_dopamine_2,warp_rm}, or elicit goal-conditioned process reasoning through reinforcement learning~\cite{primo_r1}.
Another line of work predicts alternative quantities such as temporal distance to completion~\cite{rynnvalue} or action values for the current subtask~\cite{seeq}, alongside benchmarks examining partial observations and multi-stage execution~\cite{opengvl,progresslm}.
However, relative rankings, temporal distance, and action values target distinct supervision signals, leaving a unified absolute progress scale across tasks unspecified.
This gap becomes acute in complicated long-horizon manipulation, where subtasks advance at uneven rates and later failures can revoke credit from earlier achievements.
\textbf{NimCritic} addresses this limitation by defining quantitative absolute progress through observable action effects within object-level subtasks.

\paragraph{Human-Aligned Reward Learning.}
Human-aligned reward learning translates human judgments of task execution into structured training supervision.
Pairwise trajectory comparisons constrain relative ordering~\cite{robometer}, while outcome rubrics evaluate partial completion and counterfactual task mismatches~\cite{roboreward}.
To obtain dense numerical targets, stage-aware methods partition demonstrations into qualitative stages and apply duration-based linear interpolation~\cite{sarm}, while recent extensions introduce reusable action primitives~\cite{sarm2} or synthesize VLM-generated subtask annotations with procedural progress budgets~\cite{procvlm}.
Complementary efforts incorporate directional judgments to distinguish advancement, stagnation, and regression~\cite{arm}, or apply rollout segmentation to isolate recovery segments~\cite{help_vlac_cut}.
Yet manual qualitative stage splitting remains subjective and struggles to quantify non-monotonic behaviors across long-horizon multi-subtask episodes.
Our \textbf{action-aware quantification} formalizes full credit, temporary attempt credit, and credit revocation within this procedural view, combining VLM pre-annotation with human verification for scalable supervision.

\paragraph{World Models.}
World models capture environment dynamics by predicting future observations or latent states, supplying informative temporal priors for control and evaluation.
Early approaches directly use video prediction likelihoods as reward signals~\cite{viper}, whereas world action models co-train video generation and action prediction while bypassing future-video synthesis at inference~\cite{yuan2026fastwam}.
Predicting future proprioception or latent dynamics similarly improves value and critic learning~\cite{viva,wcm}.
Architecturally, WVM couples video and value diffusion transformers through asymmetric attention~\cite{wvm}, while Self Forcing mitigates exposure bias in autoregressive diffusion via self-generated rollouts~\cite{self_forcing}.
Explicit links between intermediate video representations and quantitative progress provide a direct way to examine how predictive video learning supports reward estimation.
\textbf{NimCritic} bridges this gap by introducing contrastive Video--Progress Alignment alongside Historical Progress Perturbation, and systematically analyzes how joint video prediction sharpens task-object attention and synchronizes visual--progress representations.

\section{Predictive Long-Horizon Progress Modeling}
\label{sec:method}
\subsection{Architecture}
\begin{figure}[!tbp]
\centering
\includegraphics[width=\linewidth]{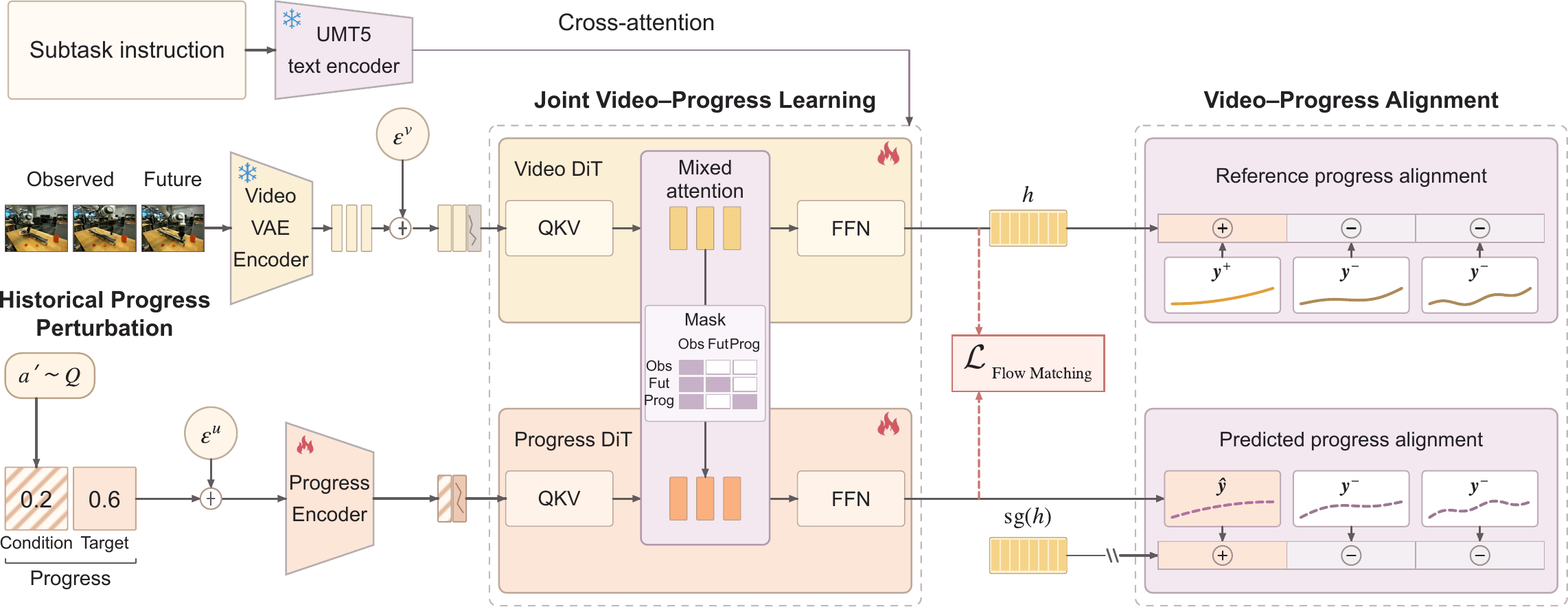}
\caption{\textbf{NimCritic training architecture:} the video expert learns visual dynamics and supplies intermediate features, the progress expert predicts progress and completion, and Video--Progress Alignment connects the intermediate video feature $h$ to reference and predicted progress trajectories using contrastive supervision.}
\label{fig:architecture}
\end{figure}

Our framework comprises the \textbf{action-aware quantification} pipeline, a VLM front-end for subtask decomposition, and a predictive critic, together forming \textbf{NimCritic} (Figure~\ref{fig:architecture}).
First, the \textbf{action-aware quantification} pipeline establishes a unified, scalable protocol that generates human-aligned dense progress and completion supervision for training and evaluation.
Second, VLM-based subtask decomposition serves as the front-end module that partitions long-horizon multi-task videos into ordered, object-level subtask segments.
Third, given each subtask goal and observed video window, the critic learns and predicts within-subtask progress and a dedicated completion output before duration-weighted global composition.

\subsection{Progress Annotation Pipeline}
Procedural task annotations provide semantic subgoals and temporal structure for progress supervision~\cite{sarm,procvlm}; our \textbf{action-aware quantification} assigns numerical credit to the actions within those subtasks.
Under this specification, our pipeline produces an 800+ hour action-aware training dataset and LOOM, a long-horizon progress benchmark of 300 trajectories and 19,618 human-calibrated evaluation frames spanning diverse embodiments and non-expert behaviors including hesitation, progress regression, and recovery.

To achieve consistent quantification, our rule-based specification distinguishes an action's physical class from its semantic task effect: the same motion can complete a subgoal, make an incomplete attempt on an object, or revoke credit from an earlier achievement.
Specifically, we define a subtask as consecutive operations on the same manipulated object and use action boundaries to locate its temporal breakpoints.
Each physical action class has a fixed weight $w_c$ estimated from dataset-wide normalized action durations, while its effect on the task determines whether an occurrence earns or revokes credit.
Within each subtask, local progress at video timestamp $t$ is the retained action credit normalized by a fixed subtask budget $B$; the subtask index is omitted below for clarity.
Let $\mathcal I_k$ contain the action occurrences in subtask $k$. Occurrence $i$ spans $[t_i^{\mathrm{s}},t_i^{\mathrm{e}}]$ ($t_i^{\mathrm e}>t_i^{\mathrm s}$), with physical class $c_i$, credit factor $\alpha_i$, and reference set $\mathcal E_i$ indexing earlier occurrences whose credit it revokes.
At timestamp $t$, its clipped elapsed fraction $\ell_i(t)$ and net action credit $\eta_i$ are defined as
\begin{equation}
\ell_i(t)=\operatorname{clip}\!\left(\frac{t-t_i^{\mathrm s}}{t_i^{\mathrm e}-t_i^{\mathrm s}},\,0,\,1\right),\qquad
\eta_i=\alpha_iw_{c_i}-\sum_{j\in\mathcal E_i\cap\mathcal I_k}\alpha_jw_{c_j}.
\label{eq:action-credit}
\end{equation}

Crucially, when a subtask ends incomplete, the VLM auto-labeler compares the final observed state against the intended goal to infer the set $\mathcal R$ of necessary remaining or restorative action occurrences.
Summing the net credits of observed actions and the class weights of the remaining actions in $\mathcal R$ defines the subtask budget $B>0$.
A forward effect receives full class credit, an engaged-only attempt receives temporary half credit, and other occurrences receive zero credit. Subsequent revocations remove the actual credit of their referenced occurrences in $\mathcal E_i$ once, yielding the absolute subtask progress $p_t$ at timestamp $t$:
\begin{equation}
p_t=\frac{\sum_i\eta_i\ell_i(t)}{B},\qquad
B=\sum_i\eta_i+\sum_{j\in\mathcal R}w_{c_j}.
\label{eq:credit}
\end{equation}

Across subtasks, each credit remains associated with the subtask that earned it. Its retained credit $r_k(t)$ accounts for subsequent revocations, including those from another subtask:
\begin{equation}
r_k(t)=\sum_{j\in\mathcal I_k}\alpha_jw_{c_j}\left[\ell_j(t)-\sum_{i:\,j\in\mathcal E_i}\ell_i(t)\right].
\label{eq:retained-credit}
\end{equation}
Each referenced credit is revoked once. These updated contributions determine the global annotation: in Figure~\ref{fig:overview}, the later push revokes credit earned in subtask I, and recovery actions earn new credit.

This formulation naturally captures both progress increases and regressions, with recovery defined as a subsequent increase following an earlier decrease, while hesitation is identified alongside its plateauing progress curve.
In our pre-annotation pipeline, a VLM determines whether each subtask succeeds, describes its intended goal, identifies subtask and action boundaries, and records the necessary remaining actions $\mathcal R$. Human verification refines temporal boundaries and the timing of temporary attempt credit and its revocation; deterministic post-processing then computes the dense progress values from these structured records.
Training annotations undergo sampled human verification, whereas every trajectory in LOOM receives full temporal verification. Reviewers adjust subtask and action timestamps, and the quantification rules determine the progress values, making the protocol scalable while reducing reliance on intuitive numerical judgments (Appendix~\ref{app:data}).

\subsection{VLM-Based Subtask Decomposition}
To extend action-aware progress estimation to long-horizon multi-task trajectories, a VLM front-end partitions the complete video into ordered, contiguous object-level subtask segments conditioned on the task instruction.
Our segmentation prompt enforces structured object-centric boundaries: a complete grasp-to-release placement stays within a single subtask, whereas acquiring a tool and using it on another object are separated into consecutive subtasks; multiple instances of an object category form distinct subtasks, while slips, failed attempts, and immediate retries on the same object instance remain consolidated within one segment so local regressions and recoveries are preserved.
Each generated subtask instruction $g_k$ specifies the intended task goal through its manipulated object and desired outcome. The VLM records the observed success or failure separately, keeping the intended goal consistent across successful executions, attempts, and retries.
Predicted segments shorter than the critic's minimum observation window are merged into an adjacent segment to form $K$ segments, after which each subtask clip is paired with its local instruction $g_k$ and initialized at zero progress at its first frame, allowing the critic to score all subtasks in parallel as self-contained short trajectories.
At inference, each independently scored segment supplies a predicted local progress curve and exit score. For timestamp $t$ in predicted segment $k$ among $K$ segments, let $n_k$ denote its frame count and $\widehat e_k$ its predicted exit progress; duration-weighted global prediction is
\begin{equation}
\widehat p_t^{\mathrm{global}}=\frac{\sum_{k^\prime<k}n_{k^\prime}\widehat e_{k^\prime}+n_k\widehat p_t^{(k)}}{\sum_{k^\prime=1}^{K}n_{k^\prime}}.
\label{eq:hierarchy}
\end{equation}
Each preceding subtask $k^\prime<k$ retains its predicted exit contribution $n_{k^\prime}\widehat e_{k^\prime}$ while later subtasks reserve duration weight in the denominator, and overall task completion is obtained by scoring the full video trajectory against the original task instruction.

\subsection{Predictive Critic}
\paragraph{Joint Video--Progress Learning.}
To enable a shared visual representation to capture both how physical states evolve and how those transitions advance the task goal, we design Joint Video--Progress Learning by adapting the dual-stream architecture of WVM~\cite{wvm} to jointly predict video latents and action-aware progress targets (Figure~\ref{fig:architecture}).
A frozen video VAE encoder $\operatorname{Enc}$ maps observed and future video frames ($o_{\leq t}$ and $o_{>t}$) into video latents, while a progress encoder maps progress--completion inputs into latent tokens; coupled video and progress diffusion transformers~\cite{dit} with trainable parameters $\theta$, initialized from the Wan family~\cite{wan} with cross-attention conditioning from a frozen UMT5 text encoder~\cite{umt5,t5}, process the video stream ($\mathrm v$) and the progress--completion stream ($\mathrm u$) conditioned on subtask instruction $g_k$.
Through asymmetric mixed attention, observed-video tokens attend causally within the observed prefix while the progress expert attends jointly to observed-video features and progress tokens to decode dense trajectories.
For modality $m\in\{\mathrm v,\mathrm u\}$ with clean latent sequence $z^m$ and noise level $\sigma_m$, we draw standard Gaussian noise $\epsilon^m$ ($\epsilon^{\mathrm v}$ and $\epsilon^{\mathrm u}$ in Figure~\ref{fig:architecture}), form the corrupted latent $z_{\sigma_m}^m=(1-\sigma_m)z^m+\sigma_m\epsilon^m$, and train the modality-specific vector field $f_\theta^m$ under loss weights $\lambda_m(\sigma_m)$ by masked flow matching~\cite{flow_matching}:
\begin{equation}
\mathcal L_{\mathrm{FM}}=\mathbb E\!\left[
\sum_{m\in\{\mathrm v,\mathrm u\}}\lambda_m(\sigma_m)
\left\|f_\theta^m(z_{\sigma_m}^m;\sigma_m,\mathcal C)-(\epsilon^m-z^m)\right\|_{\Omega_m}^{2}\right].
\label{eq:flow}
\end{equation}
Here, $\mathbb E$ denotes expectation over training examples and noise samples, conditioning $\mathcal C$ comprises the subtask instruction $g_k$, observed video $o_{\leq t}$, and historical progress anchor $a$, and $\|\cdot\|_{\Omega_m}^{2}$ computes the mean squared residual over valid supervised coordinates selected by mask $\Omega_m$.
Future video observations $o_{>t}$ provide predictive training supervision. At inference, the critic encodes observed frames once and reuses cached key--value states throughout progress--completion denoising.

\paragraph{Historical Progress Perturbation.}
During sequential inference, \textbf{NimCritic} conditions each sliding window on its own prior progress prediction, which may deviate from the clean anchors available during training.
To improve robustness against upstream estimation errors, we introduce Historical Progress Perturbation, perturbing the historical anchor by bounded random replacement while recomputing relative targets from unchanged absolute annotations.
For a historical anchor $a$ and absolute target $p_t$, we define the relative encoding function $d(a,p_t)$ to obtain the anchor-normalized progress delta $\delta=d(a,p_t)$ and the inverse reconstruction function $D(a,\delta)$ as
\begin{equation}
d(a,p_t)=\begin{cases}\dfrac{p_t-a}{1-a},&p_t\geq a,\\[4pt]\dfrac{p_t-a}{a},&p_t<a,\end{cases}\qquad
D(a,\delta)=\begin{cases}a+\delta(1-a),&\delta\geq 0,\\[2pt]a(1+\delta),&\delta<0,\end{cases}
\label{eq:hpp}
\end{equation}
with $d(1,1)=0$ and numerically safeguarded boundary denominators; at inference, $D(a,\widehat\delta_t)$ recovers the predicted subtask progress $\widehat p_t=D(a,\widehat\delta_t)$ from the predicted relative increment $\widehat\delta_t$.
During training, we sample a replacement anchor $a^\prime\sim Q$ from a bounded proposal distribution $Q$ and recompute the relative target $\delta^\prime=d(a^\prime,p_t)$, preserving $D(a^\prime,\delta^\prime)=p_t$ so the critic learns to recover accurate progress from perturbed anchors.

\paragraph{Video--Progress Alignment.}
Because asymmetric mixed attention is unidirectional, we introduce Video--Progress Alignment so the video branch's intermediate observed-video feature $h$ is explicitly aligned with both the ground-truth progress trajectory and the critic's own predicted trajectory (Figure~\ref{fig:architecture}).
The Video--Progress Alignment loss $\mathcal L_{\mathrm{VPA}}$ teaches video features to distinguish annotated curves from perturbed alternatives and trains denoised progress curves to agree with those features.
Let $h$ be the projected, normalized intermediate feature extracted from observed-video tokens, $\mathbf y$ a candidate progress--completion curve in scored latent coordinates, $\mathbf y^+$ the annotated ground-truth curve, and $\mathbf y^-\sim q(\mathbf y^-\mid\mathbf y^+)$ a perturbed negative curve sampled from conditional proposal $q(\mathbf y^-\mid\mathbf y^+)$ that varies progress while preserving completion coordinates, scored by similarity function $S(h,\mathbf y)$.
Denoting by $\mathrm{pred}$ the restriction to unconditioned progress--completion positions, with noisy latent suffix $z^{\mathrm u}_{\sigma_{\mathrm u},\mathrm{pred}}$ and predicted vector-field suffix $f^{\mathrm u}_{\theta,\mathrm{pred}}$, the progress branch forms the predicted clean curve $\widehat{\mathbf y}$:
\begin{equation}
\widehat{\mathbf y}=z^{\mathrm u}_{\sigma_{\mathrm u},\mathrm{pred}}-\sigma_{\mathrm u}f^{\mathrm u}_{\theta,\mathrm{pred}}(z^{\mathrm u}_{\sigma_{\mathrm u}};\sigma_{\mathrm u},\mathcal C).
\label{eq:clean-curve}
\end{equation}
Combining conditional noise-contrastive estimation~\cite{conditional_nce} for reference-progress alignment with stop-gradient ($\operatorname{sg}$) predicted-progress alignment and logistic loss $\operatorname{softplus}(x)=\log(1+\exp x)$ yields the alignment loss:
\begin{equation}
\begin{aligned}
\mathcal L_{\mathrm{VPA}}=\frac12\mathbb E\Bigg[
&\operatorname{softplus}\!\left(-S(h,\mathbf y^+)+S(h,\mathbf y^-)-\log\frac{q(\mathbf y^-\mid\mathbf y^+)}{q(\mathbf y^+\mid\mathbf y^-)}\right)\\
+{}&\operatorname{softplus}\!\left(-S(\operatorname{sg}(h),\widehat{\mathbf y})+S(\operatorname{sg}(h),\operatorname{sg}(\mathbf y^-))\right)\Bigg].
\end{aligned}
\label{eq:alignment}
\end{equation}
In the second term, stop-gradient fixes the video feature and negative curve while gradients through $\widehat{\mathbf y}$ update the progress expert; both terms reuse existing features, and training minimizes $\mathcal L_{\mathrm{FM}}$ plus a weighted $\mathcal L_{\mathrm{VPA}}$.

\section{Experiments}
\label{sec:experiments}
\subsection{Experimental Setup}
\label{sec:setup}
We evaluate all methods on the same trajectory groups using each baseline's native inference pipeline, observation interface, and output definition; single-task episodes contain one subtask and multi-task episodes span multiple subtasks.
We use each baseline's original progress output, with minimal adaptation for methods that predict progress changes or temporal distance: predicted changes reconstruct the progress curve, and temporal-distance scores supply the ordering used for rank correlation.
\textbf{NimCritic} is trained for 128,000 updates on our 800+ hour dataset using 48 PPU-ZW810E accelerators with an effective batch size of 192 (four examples per accelerator and one gradient-accumulation step), using progress-only inference in both temporal directions.
\textbf{NimCritic} uses Gemini 3.8 Flash for subtask decomposition. Appendix~\ref{app:comparisons} compares the decomposition quality of three VLMs and evaluates all critics using the same cached Gemini segments.
Binary reference labels in dataset construction and evaluation are positive when annotated absolute progress exceeds $0.99$; \textbf{NimCritic} predicts success through its separately supervised completion output.
We report forward and time-reversed progress rank correlations ($\mathrm{PRC}^+$ and $\mathrm{PRC}^-$, with reverse predictions aligned to the original time axis), their harmonic combination ($\mathrm{PRC\text{-}F1}$), and terminal success accuracy ($\mathrm{TSA}$). The two temporal directions assess whether an estimator follows task advancement and regression from the same visual execution, while TSA assesses the dedicated completion decision (Appendix~\ref{app:protocol}).

\subsection{Long-Horizon Progress Estimation on LOOM}
Table~\ref{tab:benchmark} reports overall, single-task, and multi-task progress estimation on LOOM, where \textbf{NimCritic} achieves the highest overall scores across all four metrics with $\mathrm{PRC}^+$ of 0.87, $\mathrm{PRC}^-$ of 0.86, $\mathrm{PRC\text{-}F1}$ of 0.86, and $\mathrm{TSA}$ of 0.92. Its multi-task subset contains 129 trajectories and 13,131 evaluation frames, averaging 3.31 annotated subtasks and 50.73 seconds per trajectory.
\WRMTableBenchmark
Bold values indicate the best results within each baseline group. GVL reuses its per-image predictions across both temporal evaluation directions, yielding identical forward PRC, reverse PRC, and PRC-F1 scores.

On single-task episodes, \textbf{NimCritic} leads all specialized reward models in $\mathrm{PRC}^-$ (0.64), $\mathrm{PRC\text{-}F1}$ (0.67), and $\mathrm{TSA}$ (0.88).
On multi-task trajectories, \textbf{NimCritic} reaches 0.95 $\mathrm{PRC}^+$, 0.95 $\mathrm{PRC}^-$, 0.95 $\mathrm{PRC\text{-}F1}$, and 0.93 $\mathrm{TSA}$, supporting progress estimation through individual subtask goals and duration-weighted composition. For comparison, RoboDopamine-GRM-2 8B obtains 0.42 $\mathrm{PRC\text{-}F1}$ and Robometer 4B obtains 0.00 on the same multi-task subset.
Negative reverse PRC indicates ordering opposite to the aligned reference; zero PRC-F1 follows from a zero directional correlation or a sign disagreement between the two directional correlations (Appendix~\ref{app:protocol}).
The reverse correlations of 0.86 overall and 0.95 on multi-task trajectories support \textbf{NimCritic}'s sensitivity to task-relevant visual transitions, alongside the component analysis in Section~\ref{sec:ablation-components}.

General-purpose VLMs prompted with GVL (Gemini 3.8 Flash and GPT-6-Sol) achieve 0.73 and 0.68 overall $\mathrm{PRC\text{-}F1}$, while \textbf{NimCritic} surpasses both backbones in multi-task $\mathrm{PRC\text{-}F1}$ (0.95 vs.\ 0.74 and 0.68) and overall $\mathrm{TSA}$ (0.92 vs.\ 0.60 and 0.65).
Appendix~\ref{app:comparisons} further evaluates all methods under shared Gemini 3.8 Flash subtask decomposition (Table~\ref{tab:common-segmentation}) and compares \textbf{NimCritic} with direct whole-video scoring (Table~\ref{tab:direct-vs-full}): under shared decomposition, \textbf{NimCritic} leads all specialized reward models across all four metrics (0.95 vs.\ 0.86 $\mathrm{PRC\text{-}F1}$ for RynnValue 4B) and achieves the highest completion accuracy overall (0.93 vs.\ 0.65 $\mathrm{TSA}$ for GPT-6-Sol), while direct scoring with the same critic matches or exceeds the strongest external specialized baseline on seven of the twelve metrics, including $\mathrm{PRC\text{-}F1}$ across all three evaluation groups.

\FloatBarrier
\begin{figure}[H]
\centering
\includegraphics[width=\linewidth]{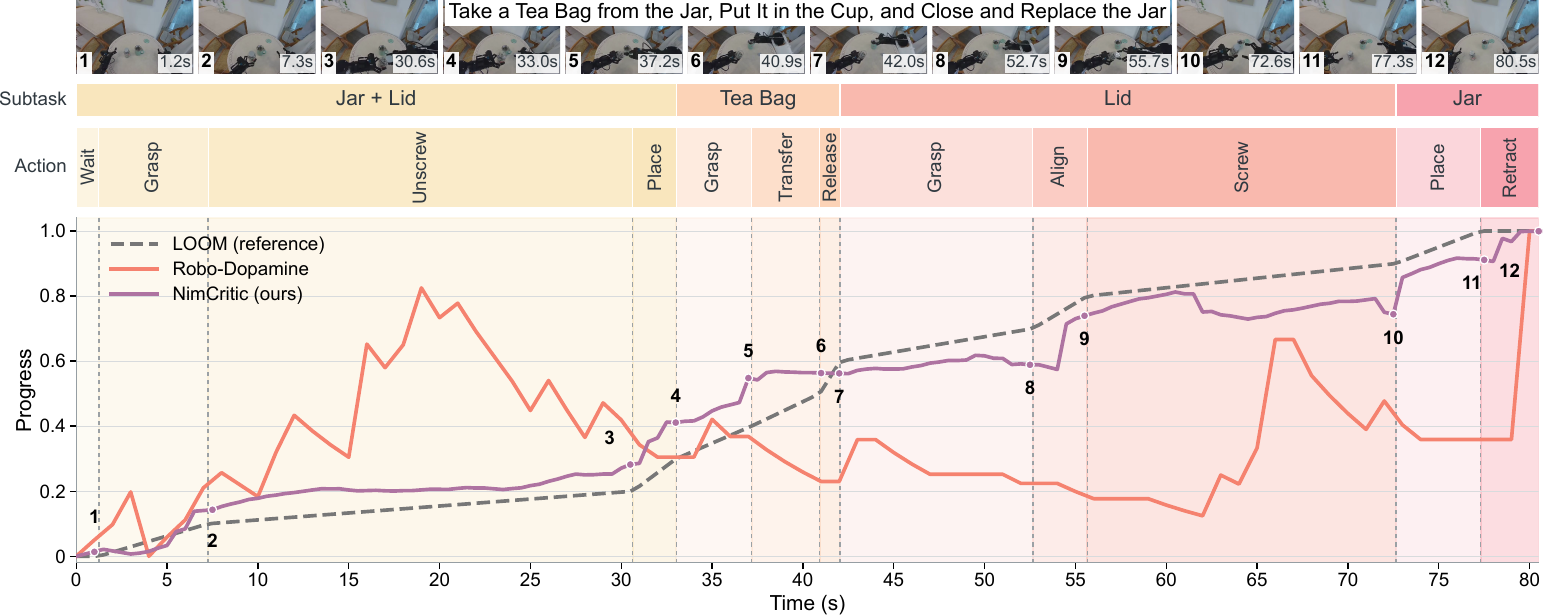}
\caption{\textbf{Progress estimation on LOOM:} \textbf{NimCritic} follows different rates of advancement across subtasks, with numbered frames marking action boundaries.}
\label{fig:long-horizon}
\end{figure}

\subsection{Progress Estimation on the Public VPB Benchmark}
\label{sec:vpb}
Table~\ref{tab:vpb} evaluates progress estimation on the four official splits of the established Video Progress Benchmark (VPB), separating seen and unseen tasks across expert and non-expert rollouts; non-expert trajectories contain at least one annotated progress regression. Zero PRC-F1 entries, including those for VLAC-Critic 2B, follow the directional-correlation rule described with Table~\ref{tab:benchmark}.
\WRMTableVPB

Among specialized reward models, \textbf{NimCritic} obtains the highest $\mathrm{PRC\text{-}F1}$ and $\mathrm{TSA}$ in all four VPB splits, reaching 0.77 overall $\mathrm{PRC\text{-}F1}$ and 0.82 $\mathrm{TSA}$ (0.88 and 0.83 $\mathrm{PRC\text{-}F1}$ on expert seen and unseen tasks; 0.66 and 0.67 on non-expert seen and unseen tasks).
The unseen splits yield 0.84 and 0.78 $\mathrm{TSA}$ on Expert Unseen and Non-expert Unseen, respectively, showing transfer across manipulation tasks; the non-expert results also cover execution with progress regressions (Section~\ref{sec:ablation-data}).

Compared with general-purpose VLMs using GVL (0.73 $\mathrm{PRC\text{-}F1}$ and 0.50 $\mathrm{TSA}$ for GPT-6-Sol; 0.73 and 0.47 for Gemini 3.8 Flash), \textbf{NimCritic} improves overall $\mathrm{PRC\text{-}F1}$ to 0.77 and surpasses GVL (GPT-6-Sol) by about 65\% in $\mathrm{TSA}$ (0.82 vs.\ 0.50), demonstrating the value of jointly training action-aware progress trajectories and a dedicated completion output.

\begin{figure}[H]
\centering
\includegraphics[width=\linewidth]{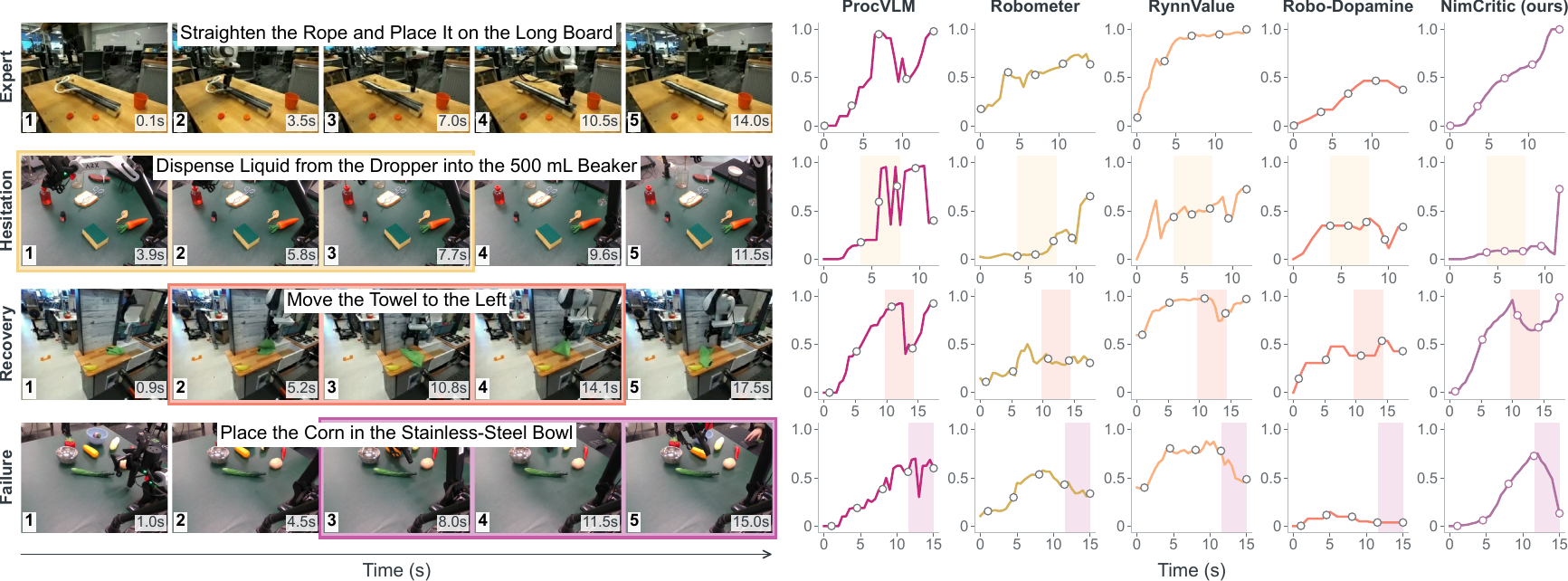}
\caption{\textbf{Progress estimation on VPB across hesitation, recovery, expert execution, and failure.}}
\label{fig:single-task}
\end{figure}

Figure~\ref{fig:single-task} compares \textbf{NimCritic} with representative reward-model baselines across these execution patterns, with five hollow markers per curve linking predictions to the displayed observations.

\subsection{Effect of Training Data Composition}
\label{sec:ablation-data}
Table~\ref{tab:data-study} evaluates different training data compositions on VPB. Every configuration includes Bridge, Fractal/RT-1, and reannotated trajectories from ARX, DROID, VLABench, and LIBERO. The three columns track additions of AgiBot, egocentric human video from EgoDex and SynData, and trajectories labeled non-expert in the ARX/DROID/VLABench/LIBERO mixture.
\WRMTableData

Adding egocentric human manipulation video raises $\mathrm{PRC}^-$ from 0.42 to 0.72, $\mathrm{PRC\text{-}F1}$ from 0.50 to 0.69, and $\mathrm{TSA}$ from 0.42 to 0.86.
Adding non-expert robot trajectories raises $\mathrm{PRC}^+$ from 0.62 to 0.72, $\mathrm{PRC\text{-}F1}$ to 0.63, and $\mathrm{TSA}$ to 0.88. AgiBot is tracked separately because it adds real-robot demonstrations with multiple subtasks, extending the training mixture's coverage of long-horizon execution; the full mixture yields 0.73 $\mathrm{PRC}^+$, 0.81 $\mathrm{PRC}^-$, and 0.77 $\mathrm{PRC\text{-}F1}$ alongside 0.82 $\mathrm{TSA}$.

\Needspace{8\baselineskip}
\subsection{Ablation Study}
\label{sec:ablation-components}
\paragraph{Results Analysis.}
Table~\ref{tab:components} evaluates how Joint Video--Progress Learning, Historical Progress Perturbation, and Video--Progress Alignment contribute to overall progress estimation on VPB.
\WRMTableComponents

Starting from a frozen video backbone (0.35 $\mathrm{PRC\text{-}F1}$ and 0.24 $\mathrm{PRC}^-$), Joint Video--Progress Learning raises $\mathrm{PRC\text{-}F1}$ to 0.59 and $\mathrm{PRC}^-$ to 0.53, providing the largest PRC-F1 gain in this comparison.
Adding Historical Progress Perturbation yields 0.74 $\mathrm{PRC}^+$, 0.67 $\mathrm{PRC\text{-}F1}$, and 0.87 $\mathrm{TSA}$; adding Video--Progress Alignment yields 0.81 $\mathrm{PRC}^-$ and 0.77 overall $\mathrm{PRC\text{-}F1}$ alongside 0.82 $\mathrm{TSA}$.

\begin{figure}[!htbp]
\centering
\includegraphics[width=\linewidth]{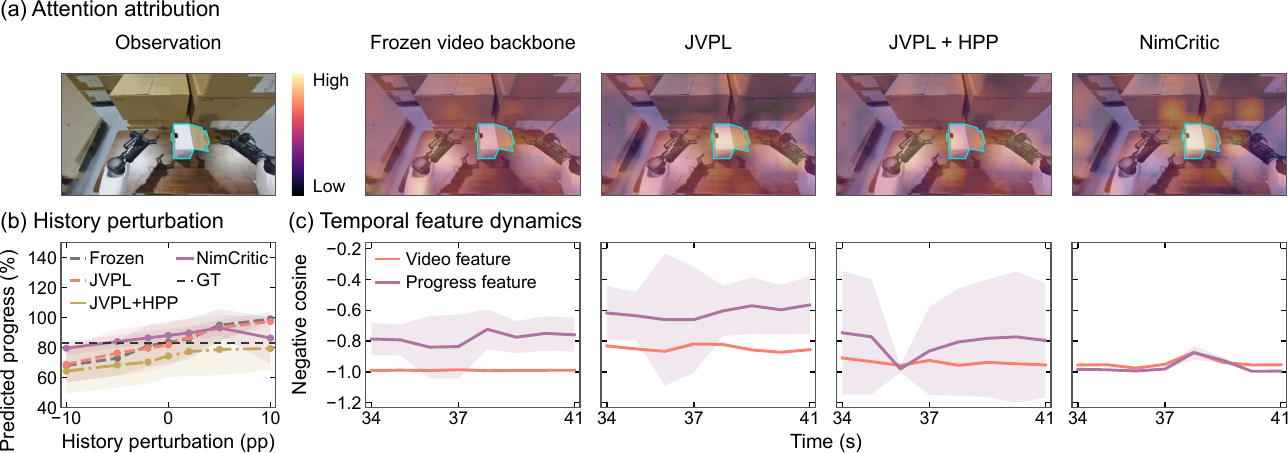}
\caption{\textbf{Mechanism analysis on LOOM:} the carton-closing example shows (a) task-object attribution, (b) responses to perturbed progress history, and (c) video and progress feature profiles, with shading denoting one sample standard deviation across five denoising runs.}
\label{fig:mechanism}
\end{figure}

\paragraph{Mechanism Analysis.}
Figure~\ref{fig:mechanism} examines visual attribution, historical sensitivity, and cross-modal alignment on a carton-closing trajectory from LOOM (AgiBotWorld).
In Figure~\ref{fig:mechanism}(a), the target-object attribution share rises from 6.43\% for the frozen backbone to 17.71\% for \texttt{JVPL} and 23.32\% for \textbf{NimCritic}, suggesting that predictive video learning helps the critic use visual evidence from the manipulated object.
In Figure~\ref{fig:mechanism}(b), Historical Progress Perturbation reduces sensitivity to the supplied history: mean prediction shifts are 5.48 percentage points for \texttt{JVPL+HPP} and 5.50 for \textbf{NimCritic}, compared with 8.84 for \texttt{JVPL}. With a $-10$ percentage-point history offset, \textbf{NimCritic} predicts 79.46\% against the 82.69\% reference.
In Figure~\ref{fig:mechanism}(c), \textbf{NimCritic}'s video and progress feature profiles evolve synchronously (Pearson $r=0.93$), consistent with the correspondence encouraged by Video--Progress Alignment.

\FloatBarrier
\subsection{Successful-Rollout Retrieval for Policy Evaluation}
Beyond dense progress estimation, recovering human-selected successful rollouts supports policy evaluation and data selection by identifying executions that satisfy the task goal.
Table~\ref{tab:policy} evaluates this capability on the large-scale RoboArena benchmark~\cite{roboarena}. Recall ($\mathrm R$) measures the fraction of human-selected successful rollouts recovered, and hit rate ($\mathrm{HR}$) measures the fraction of positive sessions with at least one recovered success; scores are reported as percentages.
\WRMTablePolicy

\textbf{NimCritic} obtains 96.72\% overall recall and 98.09\% hit rate, with 98.08\%, 85.71\%, and 95.28\% recall in the two-, four-, and seven-policy groups, outperforming RoboDopamine-GRM-2 8B (90.66\% recall and 93.66\% HR).
Conservative endpoint progress estimates can lead to cautious success decisions even on completed rollouts, offering one explanation for the zero recall of VLAC-Critic variants in the four-policy group. Separately supervised completion gives \textbf{NimCritic} a dedicated output for recognizing goal satisfaction alongside continuous task advancement.
Together, the progress and completion outputs support dense execution analysis and retrieval of human-selected successful rollouts for policy evaluation.

\Needspace{6\baselineskip}
\section{Conclusion}
We have presented \textbf{action-aware quantification} and \textbf{NimCritic} for unified, human-aligned long-horizon progress modeling in robotic manipulation.
Object-level subtasks and weighted action credits give task progress an explicit numerical meaning across attempts, regressions, and recoveries, enabling scalable VLM pre-annotation with human verification and the human-calibrated LOOM benchmark.
Coupling VLM-based subtask decomposition with Joint Video--Progress Learning, Historical Progress Perturbation, and Video--Progress Alignment grounds progress predictions in predictive visual dynamics, supporting accurate long-horizon progress estimation and successful-rollout retrieval for policy evaluation.
Together, these contributions establish a scalable foundation for dense progress feedback across downstream robotic manipulation applications.

\clearpage
\begingroup
\small
\setlength{\bibsep}{3.6pt}
\urlstyle{same}
\bibliographystyle{plainnat}
\bibliography{references}
\endgroup
\clearpage
\appendix
\setcounter{table}{0}
\renewcommand{\thetable}{A\arabic{table}}
\renewcommand{\theHtable}{A\arabic{table}}

\section{Implementation Details}
\label{app:details}
\subsection{Notation}
\label{app:notation}
Table~\ref{tab:notation} summarizes the mathematical symbols and operators used in the progress specification and model.
\begin{table}[!htbp]
\centering
\fontsize{8.0}{9.1}\selectfont
\renewcommand{\arraystretch}{0.96}
\begin{tabularx}{\linewidth}{@{}lX@{}}
\toprule
Symbol & Definition \\
\midrule
$t$ & Video timestamp. \\
$i,j$ & Indices of action occurrences, including revoked and remaining occurrences. \\
$t_i^{\mathrm s},t_i^{\mathrm e}$ & Start and end timestamps of action occurrence $i$, respectively. \\
$c_i$ & Physical action class of occurrence $i$. \\
$w_c$ & Fixed weight of physical action class $c$. \\
$\alpha_i$ & Positive credit factor of occurrence $i$: full, half, or zero credit. \\
$\mathcal E_i$ & Earlier occurrences whose credit occurrence $i$ revokes, including across subtasks. \\
$\mathcal I_k$ & Action occurrences belonging to subtask $k$. \\
$\mathcal R$ & Necessary remaining action occurrences inferred for incomplete subtasks. \\
$\ell_i(t)$ & Clipped fraction of occurrence $i$ elapsed at time $t$. \\
$\eta_i$ & Net credit of occurrence $i$, after its revocations. \\
$B$ & Fixed budget of the subtask, including its required remaining actions. \\
$p_t,\widehat p_t$ & Annotated and predicted absolute progress at time $t$ within a subtask. \\
$k,k^\prime$ & Indices of segments in the annotated or predicted partition. \\
$K$ & Segment count in the annotated or predicted partition; predicted short spans are merged. \\
$n_k$ & Predicted segment frame count, used as its duration weight. \\
$r_k(t)$ & Retained action credit of subtask $k$ at $t$, including later revocations. \\

$\widehat e_k$ & Predicted progress at the exit of segment $k$. \\
$\widehat p_t^{(k)}$ & Predicted progress within semantic subtask $k$. \\
$\widehat p_t^{\mathrm{global}}$ & Global prediction reconstructed by the duration-weighted subtask composer. \\
$m\in\{\mathrm v,\mathrm u\}$ & Modality: video ($\mathrm v$) or progress--completion ($\mathrm u$). \\
$z^m$ & Clean latent sequence for modality $m$. \\
$\epsilon^m$ & Gaussian noise for modality $m$. \\
$\sigma_m$ & Noise level of modality $m$. \\
$z^m_{\sigma_m}$ & Corrupted modality latent at noise level $\sigma_m$. \\
$g_k$ & Instruction specifying the goal of subtask $k$. \\
$o_{\leq t}$ & Video observations available through the indicated scoring time. \\
$o_{>t}$ & Future video observations used for predictive training supervision. \\
$\operatorname{Enc}$ & Frozen video encoder that produces latent observations. \\
$\theta$ & Trainable model parameters. \\
$f_\theta^m$ & Predicted modality-specific vector field. \\
$\mathcal C$ & Instruction, observed-video, and historical-anchor conditioning. \\
$\Omega_m$ & Mask selecting valid supervised coordinates for modality $m$. \\
$\lambda_m(\sigma_m)$ & Modality/noise-level flow-loss weight, including scheduler weighting. \\
$\mathcal L_{\mathrm{FM}}$ & Masked joint flow-matching objective. \\
$a,a^\prime$ & Original and randomly replaced historical progress anchors, respectively. \\
$d(a,p_t)$ & Function encoding absolute progress relative to anchor $a$. \\
$\delta,\delta^\prime,\widehat\delta_t$ & Relative targets $d(a,p_t)$, $d(a^\prime,p_t)$, and predicted relative delta. \\
$D(a,\delta)$ & Inverse function reconstructing absolute progress. \\
$Q$ & Bounded proposal distribution for replacement anchors. \\
$h$ & Projected, normalized intermediate observed-video feature. \\
$\mathbf y$ & A progress--completion curve in the model's scored latent coordinates. \\
$\mathbf y^+,\mathbf y^-$ & Annotated positive and sampled negative curves, respectively. \\
$\widehat{\mathbf y}$ & Sampled-time clean estimate of the unconditioned curve suffix. \\
$\mathrm{pred}$ & Restriction to unconditioned progress--completion positions. \\
$z^{\mathrm u}_{\sigma_{\mathrm u},\mathrm{pred}}$ & Unconditioned suffix of the corrupted progress--completion latent. \\
$f^{\mathrm u}_{\theta,\mathrm{pred}}$ & Corresponding suffix of the progress vector-field prediction. \\
$q(\mathbf y^-\mid\mathbf y^+)$ & Conditional proposal probability for negative curves. \\
$S(h,\mathbf y)$ & Similarity of the video and curve embeddings. \\
$\mathcal L_{\mathrm{VPA}}$ & Video--Progress Alignment loss, averaging grounding and output alignment. \\
$\mathbb E$ & Expectation over training examples and the indicated noise/proposal samples. \\
$\operatorname{clip}(\cdot,0,1)$ & Clipping to the unit interval. \\
$\|\cdot\|^2_{\Omega_m}$ & Mean squared residual over valid supervised coordinates. \\
$\operatorname{softplus}(x)$ & Smooth logistic loss, $\log(1+\exp x)$, for a scalar argument $x$. \\
$\operatorname{sg}$ & Stop-gradient operator. \\
\bottomrule
\end{tabularx}
\caption{\textbf{Notation for the progress specification and model.}}
\label{tab:notation}
\end{table}

\subsection{Data Composition}
\label{app:data}
Table~\ref{tab:training-mixture} summarizes our six-source training dataset totaling 253,276 episodes and 856.95 hours of manipulation video, reporting the configured training sampling weights, episode counts, and raw video hours for each source (whose hour shares are shown in Figure~\ref{fig:overview}).
Within the mixed robot-data group, the corpus combines 6,930 ARX, 12,437 DROID~\cite{droid}, 411 VLABench~\cite{vlabench}, and 354 LIBERO~\cite{libero} episodes.
\begin{table}[H]
\centering
\small
\begin{tabularx}{\linewidth}{@{}Xrrr@{}}
\toprule
Dataset(s) & Sampling weight & Episodes & Hours \\
\midrule
Bridge~\cite{bridgedata_v2} & 20.00\% & 37,692 & 70.21 \\
Fractal / RT-1~\cite{rt1} & 20.00\% & 86,100 & 346.81 \\
ARX / DROID / VLABench / LIBERO & 16.67\% & 20,132 & 93.86 \\
EgoDex~\cite{egodex} & 13.33\% & 67,043 & 171.81 \\
SynData (PsiBot) & 13.33\% & 38,059 & 117.38 \\
AgiBot~\cite{agibot_world} & 16.67\% & 4,250 & 56.87 \\
\midrule
Total & 100.00\% & 253,276 & 856.95 \\
\bottomrule
\end{tabularx}
\caption{\textbf{Training data composition.}}
\label{tab:training-mixture}
\end{table}
Table~\ref{tab:benchmark-mixture} summarizes the video-source composition of the 300 evaluation trajectories in LOOM.
The imported block contains 184 evaluation videos from ARX, DROID, and LIBERO, reannotated under our \textbf{action-aware quantification} protocol and human-verified at the subtask and action boundaries.
In total, LOOM comprises 171 single-task trajectories (6,487 human-calibrated evaluation frames) and 129 multi-task trajectories (13,131 human-calibrated evaluation frames).
\begin{table}[H]
\centering
\small
\begin{tabularx}{\linewidth}{@{}Xrr@{}}
\toprule
Dataset / collection & Trajectories & Proportion \\
\midrule
ARX & 151 & 50.33\% \\
DROID~\cite{droid} & 31 & 10.33\% \\
LIBERO~\cite{libero} & 2 & 0.67\% \\
AgiBot~\cite{agibot_world} & 57 & 19.00\% \\
AgiBotWorld Beta~\cite{agibot_world} & 22 & 7.33\% \\
Galaxea~\cite{galaxea_dynamics} & 11 & 3.67\% \\
RoboMIND~\cite{robomind} & 10 & 3.33\% \\
In-house human and robot videos & 13 & 4.33\% \\
RoboCoin~\cite{robocoin} & 2 & 0.67\% \\
Bridge~\cite{bridgedata_v2} & 1 & 0.33\% \\
\midrule
Total & 300 & 100.00\% \\
\bottomrule
\end{tabularx}
\caption{\textbf{Video-source composition of LOOM.}}
\label{tab:benchmark-mixture}
\end{table}

\subsection{Model, Training, and Inference Details}
\label{app:implementation}
We initialize the visual stream from \href{https://huggingface.co/Wan-AI/Wan2.2-TI2V-5B}{Wan2.2-TI2V-5B} and encode task and subtask instructions using the frozen \texttt{google/umt5-xxl} text encoder.
During training, each window processes nine RGB frames sampled at 2~Hz and resized to $256\times448$ resolution alongside five progress--completion latent positions, optimized in bfloat16 precision with a learning rate of $5\times10^{-5}$.
The model is trained for 128,000 updates across 48 PPU-ZW810E accelerators with four examples per device and one gradient-accumulation step, yielding an effective batch size of 192.
At inference, \textbf{NimCritic} operates in progress-only mode: each sliding window encodes five observed RGB frames into two video latent frames and denoises the five-step progress--completion trajectory over 20 Euler steps.
Sliding windows advance with a stride of two positions and aggregate overlapping predictions using inverse-distance weights; each subtask initializes its first position at zero progress and conditions subsequent windows autoregressively on the running distance-weighted average of prior predictions.
For multi-task episodes, the offline VLM front-end partitions the trajectory into contiguous object-level subtask spans, merges any span shorter than five frames into an adjacent span, and composes the resulting local subtask curves into a full-trajectory prediction via Eq.~\eqref{eq:hierarchy}.

\paragraph{General-Purpose VLM Protocol and Qualitative Visualization.}
For the general-purpose VLM baselines, GVL~\cite{gvl} prompts Gemini 3.8 Flash and GPT-6-Sol (with temperature 1 and native settings, disabling extra reasoning effort for GPT-6-Sol) to predict progress percentages for up to 30 uniformly sampled frames presented in shuffled order, then restores the predictions to the original timeline.
GVL uses these per-image predictions for both forward and reverse evaluations, yielding identical $\mathrm{PRC}^+$, $\mathrm{PRC}^-$, and $\mathrm{PRC\text{-}F1}$ scores.
In our qualitative analyses, Figures~\ref{fig:long-horizon} and~\ref{fig:single-task} display complete native prediction curves from the evaluated trajectories.
In Figure~\ref{fig:mechanism}, gradient-weighted attention attribution multiplies attention weights by their progress-output derivatives~\cite{chefer2021generic}, summing signed contributions across queries, heads, layers, and denoising steps before taking magnitude.
The target-object ROI attribution share normalizes attribution over recent observed-video tokens within a fixed task-object region over source time 34--41~s, while endpoint-relative feature profiles compute negative cosine similarity against each branch's feature at the subtask endpoint across five independent denoising runs.
History sensitivity in Figure~\ref{fig:mechanism}(b) reports the mean absolute deviation from the unperturbed prediction across six nonzero history offsets ($\pm2$, $\pm5$, and $\pm10$ percentage points) averaged across five independent denoising runs.

\section{Evaluation Metrics}
\label{app:protocol}
\subsection{Progress Estimation}
For trajectory $n$, let $\mathbf p_n$ denote the reference progress curve and $\widehat{\mathbf p}_n^\pm$ the forward and reverse-query predictions aligned to the original timeline. We compute per-trajectory Spearman rank correlation $r_n^\pm=\rho_{\mathrm S}(\widehat{\mathbf p}_n^\pm,\mathbf p_n)$ using average ranks for ties.
All methods within a table use the same trajectory-duration weights $\omega_n$ and aggregation rule, with each directional correlation averaged over its finite support $\mathcal I_\pm$. PRC-F1 is then calculated from the two aggregated directional correlations, and the same duration weights aggregate TSA.
The corpus-level forward and reverse correlations ($\mathrm{PRC}^+=r^+$ and $\mathrm{PRC}^-=r^-$) and their signed harmonic combination ($\mathrm{PRC\text{-}F1}$) are defined as
\begin{equation}
r^\pm=\frac{\sum_{n\in\mathcal I_\pm}\omega_n r_n^\pm}
{\sum_{n\in\mathcal I_\pm}\omega_n},\qquad
\operatorname{PRC\text{-}F1}=\begin{cases}
\dfrac{2r^+r^-}{r^++r^-},&r^+r^->0,\\
0,&r^+r^-\leq0.
\end{cases}
\label{eq:prc-metrics}
\end{equation}
Letting $b_n$ and $\widehat b_n$ denote the reference and predicted binary task-completion indicators with indicator function $\mathbf1[\cdot]$, duration-weighted terminal success accuracy ($\mathrm{TSA}$) is given by
\begin{equation}
\operatorname{TSA}=\frac{\sum_n\omega_n\mathbf1[\widehat b_n=b_n]}
{\sum_n\omega_n},
\label{eq:tsa}
\end{equation}
where $b_n$ follows the unified progress threshold in Section~\ref{sec:setup} and $\widehat b_n$ is the predicted completion decision.

\subsection{Policy Rollout Evaluation}
For each multi-policy evaluation session $s$ on RoboArena, let $G_s$ be the set of human-selected successful rollouts, $U_s$ the set of rollouts identified as successful by a reward model's completion readout, and $\mathcal S_+=\{s:|G_s|>0\}$ the set of positive sessions.
Trajectory-level recall ($\mathrm R$)~\cite{davis2006precisionrecall} and session-level hit rate ($\mathrm{HR}$)~\cite{tamm2021metrics} are defined as
\begin{equation}
\mathrm R=\frac{\sum_s|U_s\cap G_s|}{\sum_s|G_s|},\qquad
\mathrm{HR}=\frac{1}{|\mathcal S_+|}
\sum_{s\in\mathcal S_+}\mathbf1[|U_s\cap G_s|>0],
\label{eq:retrieval}
\end{equation}
where across 3,872 sessions and 10,750 rollouts, the positive-rollout and positive-session counts $(\sum_s|G_s|,|\mathcal S_+|)$ are $676/582$, $14/8$, and $530/246$ for the two-, four-, and seven-policy groups ($1,220/836$ overall).

\subsection{Subtask Decomposition Quality}
Let $\mathcal G$ and $\mathcal P$ denote the reference and predicted subtask segment partitions on the evaluation frame grid, where each segment is a nonempty half-open interval $I$ of frame length $|I|$ and $\operatorname{IoU}(I,J)=|I\cap J|/|I\cup J|$.
Writing $H(x,y)=2xy/(x+y)$ for $x+y>0$ (and $0$ otherwise), the directional segment-length-weighted covering scores $C_{\mathrm g}$ and $C_{\mathrm p}$ and their harmonic mean $\operatorname{SegCover\text{-}F1}$ are
\begin{equation}
\begin{aligned}
C_{\mathrm g}&=\frac{\sum_{I\in\mathcal G}|I|
\max_{J\in\mathcal P}\operatorname{IoU}(I,J)}{\sum_{I\in\mathcal G}|I|},\qquad
C_{\mathrm p}=\frac{\sum_{J\in\mathcal P}|J|
\max_{I\in\mathcal G}\operatorname{IoU}(I,J)}{\sum_{J\in\mathcal P}|J|},\\
\operatorname{SegCover\text{-}F1}&=H(C_{\mathrm g},C_{\mathrm p}).
\end{aligned}
\label{eq:seg-cover}
\end{equation}
For the one-to-one matching variant $\operatorname{SegCover\text{-}F1}_{1:1}$, let $\mathcal M_\star$ denote the optimal Hungarian assignment that maximizes the sum of matched segment IoUs:
\begin{equation}
\begin{aligned}
C_{\mathrm g}^{1:1}&=\frac{\sum_{(I,J)\in\mathcal M_\star}|I|
\operatorname{IoU}(I,J)}{\sum_{I\in\mathcal G}|I|},\quad
C_{\mathrm p}^{1:1}=\frac{\sum_{(I,J)\in\mathcal M_\star}|J|
\operatorname{IoU}(I,J)}{\sum_{J\in\mathcal P}|J|},\\
\operatorname{SegCover\text{-}F1}_{1:1}&=H(C_{\mathrm g}^{1:1},C_{\mathrm p}^{1:1}),
\end{aligned}
\label{eq:seg-cover-one-to-one}
\end{equation}
where unmatched segments contribute zero to the numerator and remain in the denominator.
For boundary accuracy, let $\mathcal B_{\mathrm g}$ and $\mathcal B_{\mathrm p}$ denote the internal reference and predicted segment boundaries, and let $\Delta\mathcal B_{\mathrm g}$ denote the spacings between consecutive reference boundaries. Under adaptive boundary tolerance $\tau$, $\operatorname{Boundary\ F1}$ is
\begin{equation}
\begin{aligned}
\tau&=0.05\!\begin{cases}
\operatorname{median}(\Delta\mathcal B_{\mathrm g}),&|\mathcal B_{\mathrm g}|\geq2,\\
\operatorname{median}\{|I|:I\in\mathcal G\},&|\mathcal B_{\mathrm g}|=1,
\end{cases}\quad
\operatorname{Boundary\ F1}=\frac{2|\mathcal M_{\mathrm b}|}
{|\mathcal B_{\mathrm g}|+|\mathcal B_{\mathrm p}|},
\end{aligned}
\label{eq:boundary-f1}
\end{equation}
where $\mathcal M_{\mathrm b}$ greedily matches boundary pairs in order of increasing temporal distance within tolerance $\tau$.
Finally, the Adjusted Rand Index ($\operatorname{ARI}$) evaluates frame-level cluster agreement corrected for chance:
\begin{equation}
\operatorname{ARI}=\frac{\operatorname{RI}-
\mathbb E_{\mathrm{perm}}[\operatorname{RI}]}
{1-\mathbb E_{\mathrm{perm}}[\operatorname{RI}]},
\label{eq:partition-ari}
\end{equation}
where $\operatorname{RI}$ is the frame-pair Rand index and $\mathbb E_{\mathrm{perm}}[\operatorname{RI}]$ is its expectation under size-preserving permutations.

\section{Additional Experimental Results}
\label{app:comparisons}
\paragraph{Comparison of VLM Subtask Segmenters.}
Table~\ref{tab:segmenters} compares the subtask-decomposition quality of GPT-6-Sol, Qwen 3.8 Max, and Gemini 3.8 Flash on LOOM's 129 multi-task trajectories, using the temporal-partition metrics in Appendix~\ref{app:protocol} and equal averaging across trajectories.
Gemini 3.8 Flash obtains the highest temporal covering (0.80 SegCover-F1 and 0.78 one-to-one SegCover-F1) and boundary localization (0.41 Boundary F1), while Qwen 3.8 Max obtains 0.70 ARI. We choose Gemini as the common decomposition frontend based on these temporal-partition metrics, then use its cached segments for \textbf{NimCritic} and all shared-segment critic comparisons.
\WRMTableSegmenters

\paragraph{Controlled Comparison Under Shared Subtask Decomposition.}
To evaluate progress estimation within subtasks under a common front-end, Table~\ref{tab:common-segmentation} supplies all methods with the same cached Gemini 3.8 Flash subtask segments and duration-weighted global composition (Eq.~\eqref{eq:hierarchy}) on the 129 multi-task episodes of LOOM. Negative correlations and zero PRC-F1 entries follow the directional-correlation rule described with Table~\ref{tab:benchmark}.
Shared subtask boundaries substantially improve external critics, lifting RynnValue 4B from $0.17$ to $0.86$ $\mathrm{PRC\text{-}F1}$ and Robometer 4B from $0.00$ to $0.79$ $\mathrm{PRC\text{-}F1}$.
Under shared decomposition, \textbf{NimCritic} leads the specialized reward models (0.95 vs.\ 0.86 PRC-F1 and 0.93 vs.\ 0.71 TSA) and obtains correlation comparable to GPT-6-Sol (0.95 vs.\ 0.96 PRC-F1), alongside higher TSA (0.93 vs.\ 0.65).
GVL and the single-frame LRM baseline reuse their per-frame predictions across both temporal evaluation directions, yielding identical $\mathrm{PRC}^+$, $\mathrm{PRC}^-$, and $\mathrm{PRC\text{-}F1}$ scores.
\WRMTableCommonSegmentation

\paragraph{Effect of Subtask Decomposition on \textbf{NimCritic}.}
Table~\ref{tab:direct-vs-full} compares direct whole-video scoring (\textbf{NimCritic w/o Subtask Decomposition}) and scoring after subtask decomposition (\textbf{NimCritic}) using the same trained critic across LOOM.
Both variants share identical single-task predictions ($0.70$, $0.64$, $0.67$, and $0.88$), while on multi-task episodes, subtask conditioning and duration-weighted composition lift $\mathrm{PRC\text{-}F1}$ from $0.55$ to $0.95$ by providing focused object-level subgoals and resetting each subtask anchor to zero.
\WRMTableDirectVsFull

\end{document}